\documentclass[12pt, a4paper,titlepage]{article}

\usepackage{arxiv}

\usepackage[english]{babel}
\usepackage[utf8x]{inputenc}
\usepackage[T1]{fontenc}
\usepackage[singlespacing]{setspace}

\usepackage[para]{footmisc}
\usepackage{array,makecell}

\usepackage{natbib}
\usepackage{subcaption}

\usepackage[toc, page, title]{appendix}

\usepackage{chngcntr}
\counterwithin{figure}{section}
\counterwithin{table}{section}

\usepackage{caption}

\DeclareCaptionLabelFormat{AppendixTablesB}{B#2}
\DeclareCaptionLabelFormat{AppendixTablesC}{C#2}

\usepackage{booktabs, ltablex, makecell}
\usepackage[dvipsnames]{xcolor}
\usepackage{xcolor}
\usepackage{framed}
\colorlet{shadecolor}{SpringGreen}
\usepackage[font=scriptsize]{caption}

\newcolumntype{C}{>{\raggedright\arraybackslash}X}
\newcolumntype{Y}{>{\raggedright\arraybackslash}X}
\usepackage{graphicx}
\usepackage{pdflscape}
\usepackage{xltabular}
\usepackage{cellspace, tabularx}
\usepackage{amsmath}
\usepackage{amssymb}
\usepackage{hyperref}
\usepackage{float}
\usepackage{adjustbox}
\usepackage{subcaption}
\usepackage{breqn}
\numberwithin{equation}{section}
\usepackage{pdfpages}

\usepackage{listings}
\usepackage{subcaption}
\usepackage[T1]{fontenc}
\usepackage{tikz}
\usetikzlibrary{arrows,calc,positioning}
\usetikzlibrary{arrows,shapes,positioning}
\usetikzlibrary{decorations.markings}
\usetikzlibrary{bayesnet}
\usetikzlibrary{arrows}
\usetikzlibrary{positioning,automata}
\usepackage{dsfont}
\usetikzlibrary{decorations.pathreplacing,angles,quotes}
\usepackage[linesnumbered,ruled]{algorithm2e}
\usepackage{algpseudocode}
\usepackage{mathtools}
\usepackage[linesnumbered,ruled]{algorithm2e}
\usepackage{makecell}
\usepackage{enumitem}
\usepackage{scalerel,stackengine}
\usepackage{chngcntr}
\usepackage[nonumberlist, acronym]{glossaries}

\newcommand{\nocontentsline}[3]{}
\newcommand{\tocless}[2]{\bgroup\let\addcontentsline=\nocontentsline#1{#2}\egroup}

\counterwithin{figure}{section}
\stackMath

\newcommand\reallywidehat[1]{%
\savestack{\tmpbox}{\stretchto{%
  \scaleto{%
    \scalerel*[\widthof{\ensuremath{#1}}]{\kern-.6pt\bigwedge\kern-.6pt}%
    {\rule[-\textheight/2]{1ex}{\textheight}}
  }{\textheight}%
}{0.5ex}}%
\stackon[1pt]{#1}{\tmpbox}%
}

\DeclarePairedDelimiterX{\infdivx}[2]{(}{)}{%
  #1\;\delimsize|\delimsize|\;#2%
}

\makeglossaries

\newacronym{CNN}{CNN}{convolutional neural network}
\newacronym{DNN}{DNN}{deep neural network}
\newacronym{FFNN}{FFNN}{feed forward neural network}
\newacronym{LSTM}{LSTM}{long short-term memory network}
\newacronym{BiLSTM}{BiLSTM}{bidirectional long short-term memory network}
\newacronym{MLP}{MLP}{multilayer perceptron}
\newacronym{SVR}{SVR}{support vector regression}
\newacronym{VARMA}{VARMA}{Vector autoregressive moving average model}
\newacronym{LeakyReLU}{LeakyReLU}{leaky rectified linear unit}
\newacronym{RMSE}{RMSE}{root mean squared error}
\newacronym{MAE}{MAE}{mean absolute error}
\newacronym{nRMSE}{nRMSE}{normalized root mean squared error}
\newacronym{MASE}{MASE}{mean absolute scaled error}
\newacronym{NN}{NN}{neural network}
\newacronym{MSVR}{MSVR}{multistep support vector regression}
\newacronym{XGBoost}{XGBoost}{eXtreme Gradient Boosting}
\newacronym{S2S reversed}{S2S reversed}{sequence-to-sequence with a decoder learning a reversed feature representation}
\newacronym{S2S context}{S2S context}{sequence-to-sequence with an encoder learning a context embedding}
\newacronym{CNN-LSTM}{CNN-LSTM}{convolutional-long short-term memory network}
\newacronym{VEST}{VEST}{automated auto-regressive feature engineering package}
\newacronym{acc95}{acc95}{95 percent accuracy score}
\newacronym{AMPds2}{AMPds2}{Almanac of Minutely Power dataset Version 2}
\newacronym{REFIT}{REFIT}{REFIT electrical load measurements dataset}
\newacronym{GREEND}{GREEND}{GREEND Electrical ENergy Dataset}
\newacronym{PecanSD}{PecanSD}{Pecan Street Data}
\newacronym{w+dt}{w+dt}{weather and date-time features}
\newacronym{HEMS}{HEMS}{home energy management systems}
\newacronym{PE}{PE}{permutation entropy}
\newacronym{wPE}{wPE}{weighted permutation entropy}
\newacronym{ls-on/ls-off}{ls-on/ls-off}{last seen on and last seen off}

\title{Multistep Multiappliance Load Prediction}

\author{\href{https://orcid.org/0000-0003-3506-4744}{\includegraphics[scale=0.06]{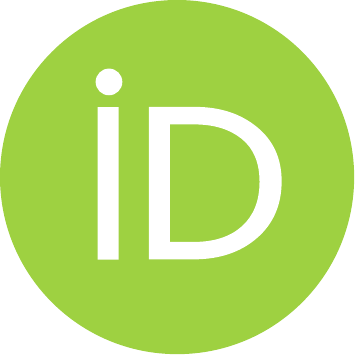}\hspace{1mm}Alona Zharova\thanks{Correspondence author.} } \\
	Humboldt-Universität zu Berlin\\
	Berlin, Germany \\
	\texttt{alona.zharova@hu-berlin.de} \\
	\And
	Antonia Scherz \\
	Humboldt-Universität zu Berlin\\
	Berlin, Germany \\
	\texttt{antonia.scherz@hu-berlin.de} \\
}

\renewcommand{\shorttitle}{Multistep Multiappliance Load Prediction}

\hypersetup{
pdftitle={titel},
pdfsubject={subject},
pdfauthor={authors},
pdfkeywords={First keyword, Second keyword, More},
}

\begin{document}
\maketitle

\begin{abstract}

A well-performing prediction model 
is vital for a recommendation system 
suggesting actions for energy-efficient consumer behavior.
However, reliable and accurate predictions depend on informative features and a suitable model design to perform well and robustly across different households and appliances.
Moreover, customers' unjustifiably high expectations to accurate predictions may discourage them from using the system in the long-term.
In this paper, we design a three-step forecasting framework to assess predictability,  engineering features, and  deep learning architectures to forecast 24 hourly load values.
First, our predictability analysis provides a tool for expectation management to cushion customers' anticipations.
Second, we design several new weather-, time- and appliance-related parameters for the modeling procedure and test their contribution to the model’s prediction performance.
Third, we examine six deep learning techniques and compare them to tree- and support vector regression benchmarks.
We develop a robust and accurate  model for the appliance-level load prediction based on four datasets from four different regions (US, UK, Austria, and Canada) with an equal set of appliances.
The empirical results show that cyclical encoding of time features and weather indicators alongside a long-short term memory (LSTM) model offer the optimal performance.

\end{abstract}


\section{Introduction}\label{introduction}

Soaring energy costs and awareness of personal responsibility to reduce carbon emissions, especially in Europe \citep{EUCommission_2021}, fuel demand for reliable and detailed energy profiling. Recommender systems that visualize upcoming costs and carbon emissions rely on energy consumption forecasts.
Thus, reliable prediction of daily appliance energy profiles with hourly consumption values facilitates consumers' and technologies' endeavors towards a sustainable everyday life.

Appliance load modeling suffers from high volatility and uncertainty in underlying data. Reliable and accurate predictions depend on informative features and a suitable model design to perform well and robustly across different households and appliances. Further, efficient prediction parameters largely depend on the time horizon and forecast granularity of the prediction problem. Deep learning addresses these challenges by efficiently processing highly variable data sources and flexibly adapting to large feature dimensions. 

Many smart home applications in the private sector struggle with interoperability, usability and satisfying consumer expectations \citep{iew_intelligent_energy}. Especially preset and high expectations to fast response time and accurate prediction outcomes demand for quick responses and transparent results of forecasters, as well as direct expectation management to cushion anticipations. Additionally, prediction frameworks should produce reliable results to support smart home- or recommendation applications. This study, therefore, designs a three-step forecasting framework assessing the predictability of underlying data, verifying potent feature groups and evaluating reliable modeling architectures for the day ahead device level load profiling in one-hour time intervals.

Until recently, existing approaches failed to simultaneously capitalize on additional information sources and efficient modeling structures. Comparative studies evaluate the predictability of appliance data leaving out an assessment of model performances on multistep forecasting tasks. Statistical approaches rely on separately modeling appliance on/off states and usage duration, while they fall short of reporting exact hourly usages. Most frameworks prove efficiency only in an aggregated setting for multiple appliances. Deep learning solutions concentrate on shorter prediction periods covering one hour ahead and higher data resolutions in minutes. Presented approaches fail to formulate longer (practical) forecasting horizons and largely ignore the potential lying in additional feature engineering.

This work addresses these shortcomings by setting up a suitable prediction problem of forecasting the next 24 hourly load values of typical appliances with different usage structures (i.e., fridge, washing machine, dryer, dishwasher, and television). The presented three-step framework evaluates data predictability, the impact of feature engineering and the performance of deep learning architectures across four different data sets from different geographical regions. Analyzed feature groups include additional information (environmental, date-time and appliance features) as well as engineered features using auto-correlation, statistical summary and phase space reconstruction techniques. Suitable model architectures includes new and existing approaches applied to related load prediction tasks. Inter alia the \gls{CNN-LSTM} and \gls{S2S reversed} architectures, as well as the \gls{MSVR} are applied to device level prediction for the first time. The code for the proposed prediction framework is available on GitHub.

The remainder of this paper is organized as follows: The subsequent subsection provides an overview of related literature. Section \ref{seq:methods} introduces the feature engineering and prediction methods as well as evaluation metrics. Section \ref{seq:exp_res} presents the datasets, preprocessing and experimental design. Section \ref{seq:results} shows the results of the different prediction methods and analyzes their implications, while their limitations and an outlook on future research follow in section \ref{seq:discussion}. Finally, section \ref{seq:conclusion} concludes the presented work.

\section{Related Literature} \label{seq:rel_literature}

Modeling aggregated residential energy consumption has been studied extensively with large success. However, few studies look at individual appliance load profiles. Despite similarities in types of prediction problems and applicable forecasting methods, individual device load forecasts are much more susceptible to uncertainty from human usage decisions. Without aggregation, as in household level (aggregated) load forecasting, random estimation errors do not cancel each other out. Additionally, individual appliances more selectively depend on influential factors such as environmental conditions or time factors. Table 2.1
, therefore, summarizes existing work on individual appliance level load predictions and applied features.

In detail, initial work modeling appliance electric profiles originates from bottom-up aggregated load profiling. \cite{capassoBottomupApproachResidential1994} analyze socioeconomic and demographic characteristics as well as average load profiles of household appliances from survey data and field measurements utilizing probability functions and a Monte Carlo extraction process. \cite{paateromodel2006} expanded this approach by formulating stochastic processes looking at collections of appliances from Finish households, separately simulating the use of each appliance.

With the advancement of data collection technologies, the most recent works build on more precise and larger datasets. Statistical approaches often indirectly estimate device-level electric consumption by modeling finite sets of operating states and the operating duration within each state to derive total load consumption profiles. Following this approach, \cite{Jin2020} utilize past on and off states for statistical analysis and probability simulations of daily television profiles. In \cite{Gao2018}'s work external environmental factors, time indicators and family internal characteristics identify similar past load patterns to resemble future patterns. A more complex approach in \cite{Yuting20} models operating states and duration times through K-means clustering combined with random sampling from state-specific probability distributions fed to a Conditional Hidden Semi-Markov model. The latter two studies predict aggregated values for groups of similar appliances. 
 
In contrast, \cite{SinghAbdulsalam18} use a more data storage-intensive approach to profiling appliances. Based on frequent appliance usage patterns stored in an incrementally updated Database Management System their Bayesian Network predicts appliance activity. To derive total load consumption profiles for various time periods (hour, day, week, etc.), they combine the predicted activity state with the average appliance electric usage and operating duration.

These indirect and empirical methods highly depend on the frequency of observations, which increases the importance of frequently observed and former values over more recent load values less frequently observed. This leads to the lagged response problem, where sudden changes in appliance usage profiles influence predictions only after a certain lag of time. In the context of user-centered applications, long adaption periods of models and predictions discourage utilization and satisfaction. 

To overcome this problem, direct approaches use time series forecasting methods to obtain the next load value from the most recent lagged time steps. In this field, research focuses on neural networks as a promising alternative capable of learning complex, nonlinear relationships between appliances and their environment. In single-step forecasting studied models include a nonlinear auto-regressive neural network predicting the next six-second time interval \citep{Laouali2022} and a linearly activated multilayer perceptron (linear regression) forecasting daily consumption \citep{Hossen2018} including weather features.

Reverting to machine learning solutions applied to appliance load forecasting \cite{lachutPredictabilityEnergyUse2014} focus on the predictability of appliance loads and compares prediction accuracies across models for single-step time horizons from one hour up to one week ahead. The authors include date-time features and achieve accuracies up to 89\% for hourly and 74\% for daily predictions. Results on predictability across different homes show large variations.

\cite{Ra20} extend the univariate single time step approach and  formulate a multivariate prediction problem by estimating load values of multiple appliances simultaneously. They design a \gls{LSTM} to predict the load values of multiple appliances for one time step ahead, comparing performances against a \gls{FFNN} and a random forest model. They additionally introduce \textit{Last_seen_on} and \textit{Last_seen_off} states as generated input features to reduce the overall errors. This multivariate prediction approach successfully captures dependencies between multiple output values of different appliances. This is an advantage over fitting separate algorithms for each output step.

In contrast and the most similar approach to the one proposed in this work, the same authors propose a framework to predict multiple future consumption values of single appliances, conditioning the model to learn dependencies among multiple time steps. \cite{Razghandi2021_1} and \cite{Razghandi2021_2} utilize an encoder-decoder type network called sequence-to-sequence model. The first variation uses long-short term memory layers, while the latter relies on bidirectional long-short term memory layers to predict load one hour ahead in 10 minute time steps.

The overview presented and outlined in Table 2.1 concludes the following: Firstly, non-deep learning approaches heavily rely on including important features for predictions while deep learning methods so far fail to utilize this additional potential. Secondly, reported deep learning approaches are limited to data granularity either smaller than 15 minutes or as aggregated as daily load values, with max forecasting horizons of six time steps resembling one hour. Assuming that individuals plan their day at least the day before and only once, practical models need larger prediction horizons covered with a more informative granularity of load estimates. Finally and regarding real-world applications, no study considers more than one data source for an equal set of appliances.

\begin{figure}
    \centering
    \label{tab:lit_review}
    \includegraphics[width=1\textwidth]{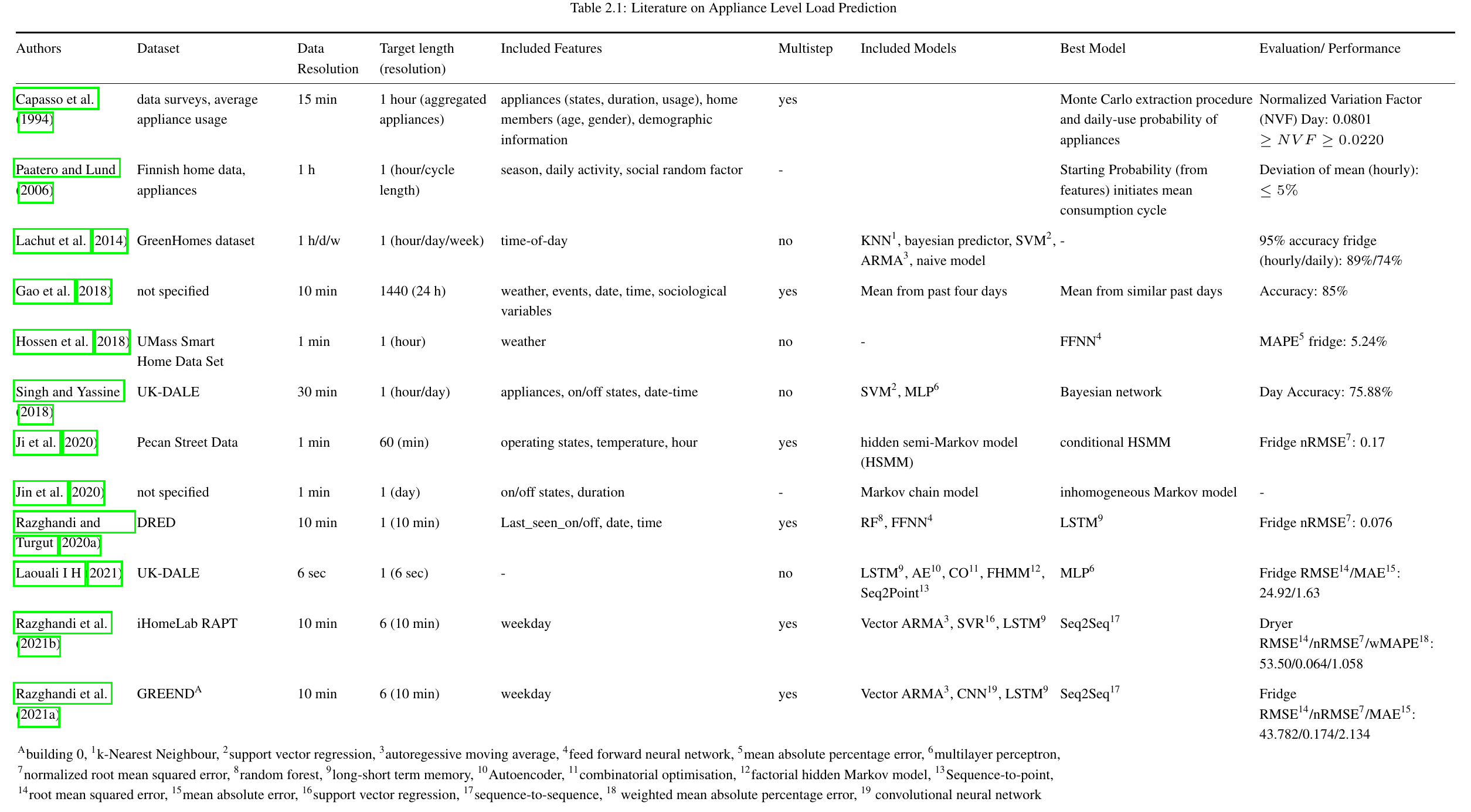}
\end{figure}

\section{Methodology} \label{seq:methods} 
This section describes a three-step framework to assess predictability, the engineered features and the deep learning architectures to forecast 24 load values. The predictability analysis forms expectations of model performance and quantifies the performance of feature groups and forecasting algorithms. The feature engineering step generates and groups similar features to distinguish the contribution of the data sources to predictions. The term features refers to variables containing either supplementary information or a reformulation of existing data (engineered features). The impact of different types of features is assessed and compared across four different datasets. The main step compares performances among different deep learning architectures and with non-deep learning (benchmark) algorithms. The following subsections introduce the methods used in this study.

\subsection{Predictability}
Predictability quantifies inherent information contained in a time series and assists in evaluating the predictive power of different forecasting methods. Model performance measures the probability of success yet it cannot provide an understanding for whether predictions improve. Intuitively, predictability estimates the highest level of performance possible for a time series and specifies whether the system is unpredictable or the model choice is poor.

A well-established measure of complexity in time series data is the \gls{PE} proposed by \cite{Bandt2002PermutationEA}. \gls{PE} captures the order relations between values and extracts a probability distribution of the ordinal patterns. \cite{AQUINO2017277} applies this \gls{PE} to electric appliance loads mapping load histograms onto a Causality Complexity-Entropy Plane to contextualize electric load behavior.\footnote{The work in \citep{AQUINO2017277} uses the REDD data.} This work relies on a modified version of the \gls{PE} the \gls{wPE} measure formulated by \cite{article13_fadlallah}. \gls{wPE} incorporates amplitude information to improve handling abrupt signal changes and more accurately assessing regular as well as noisy and (linearly) distorted data segments. The \gls{wPE} is applicable to regular, chaotic, noisy or real-world time series and fit the volatile appliance load data better than the regular \gls{PE}. \gls{wPE} uses weighted relative frequencies defined by equation \ref{eq:wpe_1} to incorporate the amplitude information into the Shannon entropy formula \ref{eq:wpe_2}. 

\begin{equation}\label{eq:wpe_1}
    p_w(\pi_i^{m, \tau}) = \frac{\sum_{j\leq N} 1_{u:\textrm{type}(u)=\pi_i}(X_{j}^{m, \tau}) * w_j} {\sum_{j\leq N} 1_{u:\textrm{type}(u) \in \Pi}(X_{j}^{m, \tau}) * w_j}
\end{equation}
\begin{equation}\label{eq:wpe_2}
    H_w(m,\tau) = - \sum_{i:\pi_{i}^{m,\tau} \in \Pi}p_w(\pi_i^{m, \tau}) \ln p_w(\pi_i^{m, \tau})
\end{equation}

The choice of weight values $w_j$ reflects a specific feature or a combination of multiple features from each vector $X_{j}^{m, \tau}$. Following \cite{article13_fadlallah} the weights are computed by the variance of each neighbors vector of $X_{j}^{m, \tau}$ in equation \ref{eq:wpe_3} with $\bar{X}_{j}^{m, \tau}$ denoting its mean (equation \ref{eq:wpe_4}).

\begin{equation}\label{eq:wpe_3}
    w_j = \frac{1}{m} \sum_{k=1}{m}(x_{j+(k-1)\tau} - \bar{X}_{j}^{m, \tau})^2
\end{equation}

with
\begin{equation}\label{eq:wpe_4}
    \bar{X}_{j}^{m, \tau} = \frac{1}{m} \sum_{k=1}{m}(x_{j+(k+1)\tau}
\end{equation}

The work in \cite{Riedly_2013} gives a guideline on how to choose parameters optimally. The final parameter sets are reported in Appendix \ref{app:wpe}.

\subsection{Feature Engineering} 
The inclusion of features and their contribution to prediction accuracy within deep learning is hardly studied in the context of appliance level load modeling. Some approaches include numerical or one-hot encoded time features without reporting their influence on predictions. Research in other areas vastly confirms large benefits from including additional information, especially in deep learning. Presumably, additional features likewise improve forecasters in the appliance load domain. The following subsections describe included feature groups proven informative in the literature on aggregated household energy consumption. In total 10 feature groups are defined: date-time, weather, appliance, \gls{ls-on/ls-off}, auto-regressive, interaction, \gls{VEST} and phase space features as well as the aggregate feature group "\gls{w+dt}" and "all" (including all features).

\subsubsection{Date-Time Features}
Opinions and practices on whether and how to include date-time features vary substantially. Common variations in load forecasting include one-hot encoded, numerical encoded and sine cosine transformed time features \citep{candanedo_data_2017}, \citep{khatoon_effects_2014}, \citep{en6031385}. Some, amongst others \cite{Razghandi2021_1}, argue against explicitly including date-time features in \gls{LSTM} modeling as the positioning of the values within the input sequence already carries the time point information. Consequently, they exclusively include a numerical representation of weekday features, but no feature with the hour and minute of the day. Though sequentially ordered input structures of \gls{LSTM} layers and numerically encoded time inputs fail to represent the cyclical nature of most date-time variables. A cyclical representation takes into account that the end and beginning of a sequence are numerically as close as time values in between, i.e. 23rd hour of the day is as close to hour zero as hour one to hour two \citep{1630255}. 

In other domains of energy forecasting, such as forecasting load at electric charging sites \citep{Unterluggauer2021} or forecasting power grid states \citep{9207536}, date-time features transformed with the sine and cosine function successfully improve prediction outcomes. Following listed examples, the hour of the day, the day of the week, the week of the month and, whenever more than a year of data is available, the month will each be represented by sine and cosine transformed values calculated as in equation \ref{eq:sine_cosine}. Further, the date-time feature group includes binary workday, holiday and weekend indicators \citep{9006868}, \citep{en11071636}.

\begin{equation}\label{eq:sine_cosine}
    \textrm{Sin}_{fe} = \sin(2\pi*n) \quad \textrm{and} \quad \textrm{Cos}_{fe} = \cos(2\pi*n)
\end{equation}
with,
\begin{equation}
 n = 
 \begin{cases}
      \textrm{hour}/24 & fe=\textrm{hour} \\
      \textrm{weekday}/7 & fe=\textrm{day of week} \\
      \textrm{monthweek}/5 & fe=\textrm{week of month} \\
      \textrm{month}/12 & fe=\textrm{month and traindata > 1 year} \\
    \end{cases}       
\end{equation}


\subsubsection{Weather Features}
Weather features have been frequently proven to improve (non-deep learning) load predictors for appliance and aggregated electricity consumption of households \citep{Sinimaa21}, \citep{Gao2018} and \citep{Hossen2018}. Weather is expected to influence appliance use as humans differ their behavior accordingly and some appliance electric consumption directly depend on environmental conditions such as temperature f.e. fridge or air conditioners. External factors influence the dependencies between load consumption and weather variables, i.e. thermostats. To assess the impact of weather features and their potential contribution to accurately predict energy usage this study includes the most commonly used weather features temperature, humidity and wind speed \citep{8905698}, \citep{MUGHEES2021114844}.

\subsubsection{Appliance Loads Features}
Especially non-deep learning approaches concentrate on the dependencies between appliances and their potential to predict single appliance load profiles. \cite{SinghAbdulsalam18} find specific appliance combinations to occur often. Intuitively specific appliances tend to be used in combination such as washing machines and dryers. However, with longer prediction horizons load values of accompanying appliances precede with a time lag, i.e. information from the washing machine one hour ago is not available for the prediction of a dryers activity.  Therefore, the predictive power of other appliance loads when modeling a specific load profile 24 hours ahead must be assessed carefully. This study integrates the load measures of selected appliances and defines their influence on a forecaster's performance. 

\subsubsection{Engineered Features}
Auto-regressive features \citep{LI2021116509} and statistical summary features such as mean and median \citep{smithMachineLearningFast2020} as well as average and standard deviation of load consumption over k-time steps  \citep{9467267} positively impact aggregated load forecasting quality. Features especially summarizing a larger past time window might especially help predict appliances used less frequently than on a daily basis. Therefore the moving average and moving max value of the past 12, 24, 36 and 72 hours compose the auto-regressive feature group. Further, four simple interaction variables between each appliance pair (sum, product, mean and standard deviation) as well as the mean and standard deviation across all appliance loads form the interaction feature group to capture more complex dependencies among appliances.

\cite{cerqueira_vest_2020} develop an automatic feature engineering package called \gls{VEST} to build auto-regressive and summary features from time series data, which covers a more extensive set of engineered features. To profit from this research and available feature engineering packages, the study uses a feature set selected by the \gls{VEST} feature engineering package for time series analysis in the \gls{VEST} feature group. Equally, the study includes \gls{ls-on/ls-off} states for the target appliance as proposed in \citep{razghandi_residential_2020}. This \gls{ls-on/ls-off} feature group useful in a multivariate prediction setting might similarly  contribute to performances in a multistep prediction setting. 

\subsubsection{Phase Space Reconstruction Features}
\cite{Drezga1998InputVS} initially used features containing reconstructed dynamics of a chaotic system as an embedding method for load forecasting. These features require phase space reconstruction\footnote{Phase space reconstruction is the foundation of nonlinear time series analysis describing the reconstruction of complete system dynamics using a single time series \citep{second_psr_def} from \citep{PSR_def_2015}.} techniques utilized for aggregated load predictions \citep{4746520}, \citep{FAN201813} and \citep{en12224349}. Several undefined exogenous factors influence the electric use of appliances directly such as human activities or environmental conditions and indirectly via factors steering human behavior. As a consequence appliance load data shows complex characteristics such as multidimensional nonlinearity and high grades of uncertainty typical for dynamic systems \citep{FAN201813}. Therefore, the motivation to transfer phase space reconstruction techniques to appliance load prediction assumes that the new phase space features reconstruct the more complex dynamics of a target appliance's usage and represent immeasurable influences on load usage profiles. 

Phase space reconstruction maps the observed time series into an embedding space that preserves the structure (topology) of the underlying dynamical system. To construct phase space reconstruction features, the target values are embedded in the space of their temporal lags using the Taken embedding \citep{takens_taken_1981}. \footnote{Takens' theorem, also named the delay embedding theorem, shows that a time series of measurements of a single observable can be used to reconstruct qualitative features of the underlying phase space system \citep{Huke2006EmbeddingND}. A phase space formulates the space in which all possible states of a system are represented. In this application, a phase space representation of the target appliance loads describes all possible target values and their complex relation to each other.} Appendix \ref{app:wpe} specifies software and parameters for the construction of the phase space features. 

\subsection{Prediction Algorithms}
A vast selection of (hybrid) neural network architectures has proven to accurately predict multistep or multivariate sequential data. Subsequent subsections describe the deep learning architectures including \gls{LSTM}, \gls{BiLSTM}, encoder-decoder networks, \gls{FFNN} and \gls{CNN-LSTM}.

\subsubsection{Long-Short Term Memory Network} \label{LSTM}
\gls{LSTM} networks, initially proposed by \cite{LSTM_basics}, are a type of recurrent neural network often used in time series prediction. They remember information from several past input steps while sequentially calculating outputs capturing the time dependencies of input variables. The architectural unique characteristics of \gls{LSTM} Networks are the \gls{LSTM} cell memory states that convey information across a chain of \gls{LSTM} cell states and update new information only if considered important. \gls{LSTM} cells process their own previous cell output $h_{t-1}$ together with new input values $x_t$ at time $t$ and update cell memory states $C_{t-1} \rightarrow C_t$ to calculate the new output value $h_t$. In this process previous cell outputs $h_{t-1}$ are filtered through input gates ${i_t}$, forget gates ${f_t}$, update gates $\widetilde{C}_t$ ($g_t$) and output gates ${o_t}$ to find the important information for updating cell memory state and calculating cell outputs. The stepwise calculations of \gls{LSTM} cell states are as follows:

\begin{equation}\label{eq:lstm1}
    f_t = \sigma_{f_t} (W_f \cdot [h_{t-1}, x_t] + b_f).
\end{equation}
\begin{equation}\label{eq:lstm2}
    i_t = \sigma_{i_t} (W_i \cdot [h_{t-1}, x_t] + b_i).
\end{equation}
\begin{equation}\label{eq:lstm3}
    \widetilde{C}_t = tanh(W_C \cdot [h_{t-1}, x_t] + b_C).
\end{equation}
\begin{equation}\label{eq:lstm4}
    C_t = f_t \cdot C_{t-1} + i_t \cdot \widetilde{C}_t.
\end{equation}
\begin{equation}\label{eq:lstm5}
    o_t = \sigma_{o_t}(W_o \cdot [h_{t-1}, x_t] + b_o).
\end{equation}
\begin{equation}\label{eq:lstm6}
    h_t = o_t*tanh(C_t).
\end{equation}

As demonstrated in Figure \ref{fig:lstm_unit}, first the input and the previous cell output are pushed through the sigmoid forget layer \ref{eq:lstm1} which determines what information from the previous cell state to keep or not to keep. Secondly, the sigmoid input layer \ref{eq:lstm2} decides which values of the cell state to update and the tanh update layer \ref{eq:lstm3} calculates candidate values for updating cell states. Equation \ref{eq:lstm4} calculates the updated cell states. Another layer, the sigmoid output layer \ref{eq:lstm5}, decides which parts of the cell state to output. The updated cell states are then pushed through a $tanh$ layer and multiplied by the output filter in Equation \ref{eq:lstm6}. This cell output together with new input values initiates the next update process of the cell state and repeats itself iteratively. This structure remembers long-term dependencies and due to the multiple gate calculations form an additive structure of the gradient term. The gradient term updates layer weights by calculating derivatives in the backward propagation process. In this context, an additive structure of the gradient term reduces the likelihood of exploding and vanishing gradients, which leads to a more stable network training making \gls{LSTM} networks a reliable choice for time series predictions \citep{LSTM_basics}.

\subsubsection{Bidirectional Long-Short Term Memory Network}
\begin{figure} 
    \centering
    \begin{minipage}[t]{0.4\textwidth}
        \centering
        \caption{\\ Functionality of the LSTM-Unit (Image Source: \cite{blog_lstm_fig})}
        \label{fig:lstm_unit}
        \includegraphics[width=0.99\textwidth]{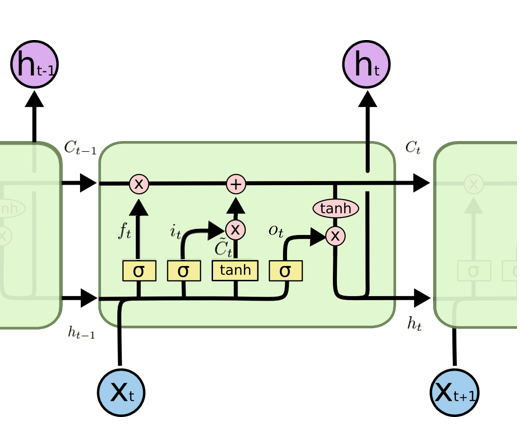}
        \break
        \tiny
        \normalsize        
    \end{minipage}\hfill
    \begin{minipage}[t]{0.6\textwidth}
        \centering
        \caption{\\ Functionality of the BiLSTM-Layer (Image Source: \cite{blog_bilstm_fig})}
        \label{fig:bilstm_layer}
        \includegraphics[width=0.99\textwidth]{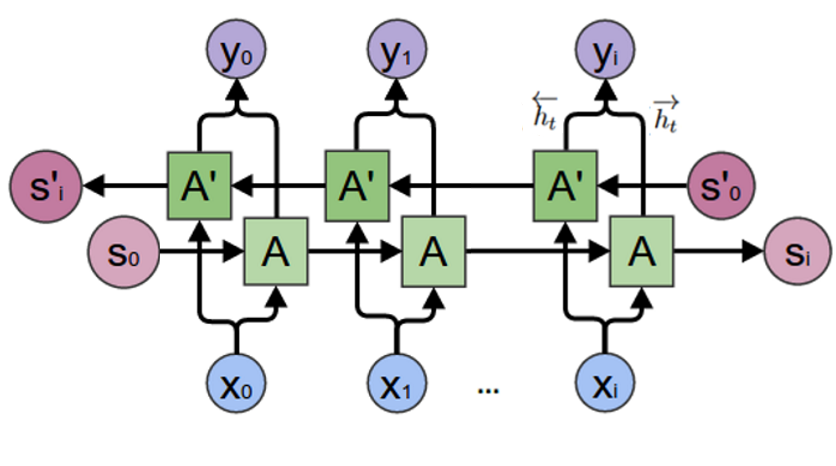} 
        \tiny
        \normalsize
    \end{minipage}
\end{figure}

\gls{LSTM} layers forward pass input values sequentially. Consequently the first output of an \gls{LSTM} cell is based solely on the first input and fails to use all values within the input sequence. A \gls{BiLSTM}, proposed by \cite{bilstm_basic}, enables all \gls{LSTM} cell states to use information from the complete input sequence, passing it forward and backward through the network (Figure \ref{fig:bilstm_layer}). To add a backward pass the \gls{BiLSTM} adapts the standard \gls{LSTM} network by adding a separated hidden layer processing sequences in reverse. The forward layer processes the input sequence from beginning to end the backward hidden layer from end to beginning. Every \gls{BiLSTM} layer contains double the number of memory cells. The information of the forward and backward pass is stored in separate hidden states ($\overleftarrow{h_t}$ and $\overrightarrow{h_t}$) which are concatenated to produce the final hidden state $h_t$. At any time step t, the forward and backward layer outputs of a \gls{BiLSTM} cell are computed using the standard \gls{LSTM} unit’s operating equations \ref{eq:lstm1}–\ref{eq:lstm5}. Then the final hidden state vector is computed by combining the hidden state sequences 
in \ref{eq:bilstm1}:

\begin{equation}\label{eq:bilstm1}
    y_t = \oplus(\overrightarrow{h_t}, \overleftarrow{h_t}).
\end{equation}

where $\oplus$ is a concatenate function. It should be noted that other operations, such as summation, multiplication, or averages, can be used instead. 
\gls{BiLSTM} Networks initially improved long-range context processing in Natural Language Processing \citep{bilstm_basic}. Momentarily, they frequently deliver state of the art performance in (load) time series predictions \citep{mugheesDeepSequenceSequence2021} and outperform deep learning architectures such as the \gls{LSTM} and \gls{CNN} \citep{article13_fadlallah}.

\subsubsection{Encoder-Decoder Networks}
Encoder-decoder models, such as the sequence-to-sequence model, likewise originate from language translation \citep{NIPS2014_a14ac55a} and prove powerful in time series forecasting \citep{8701741}, \citep{sehovac_deep_2020} and \citep{article13_fadlallah}. An encoder-decoder architecture comprises two network parts, an encoder network and a decoder network. The encoder calculates a hidden representation of the inputs encoded in hidden states and passes them on to the decoder network for calculating predictions. The difference to standard neural network training is the transfer of states instead of layer outputs between the encoder and decoder and often the two parts are pre-trained separately before the final prediction stage. This study includes two variations of the sequence-to-sequence network. The sequence-to-sequence model used by \cite{Razghandi2021_2} further called \gls{S2S reversed} and a variation of the sequence-to-sequence model proposed in the winning solution to the web traffic time series forecasting challenge introduced by \cite{blog_s2s} called \gls{S2S context}.

\begin{figure} 
    \centering
    \begin{minipage}[t]{0.5\textwidth}
        \centering
        \label{fig:ed_ra}
        \caption{\\ S2S-Model with Reversed Sequence Decoder (Image Source: \cite{Razghandi2021_2})}
        \includegraphics[width=0.99\textwidth]{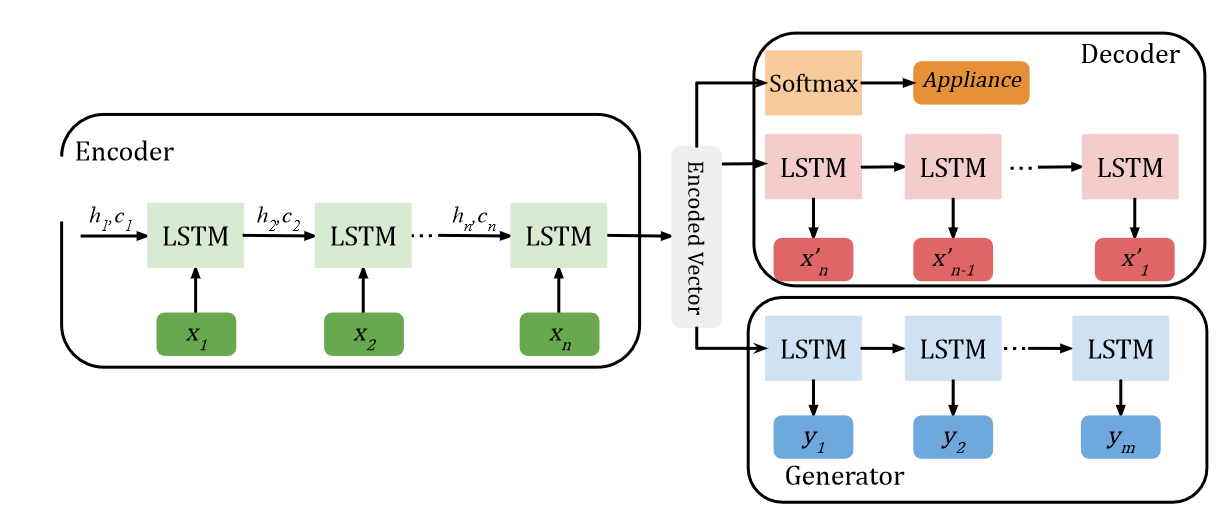}
        \break
        \break
        \tiny
        \normalsize
    \end{minipage}\hfill
    \begin{minipage}[t]{0.5\textwidth}
        \centering
        \label{fig:cnn_layer}
        \caption{\\ Functionality of 1D-CNN-Layer (Image Source: \cite{blog_lstm_fig})}
        \includegraphics[width=0.99\textwidth]{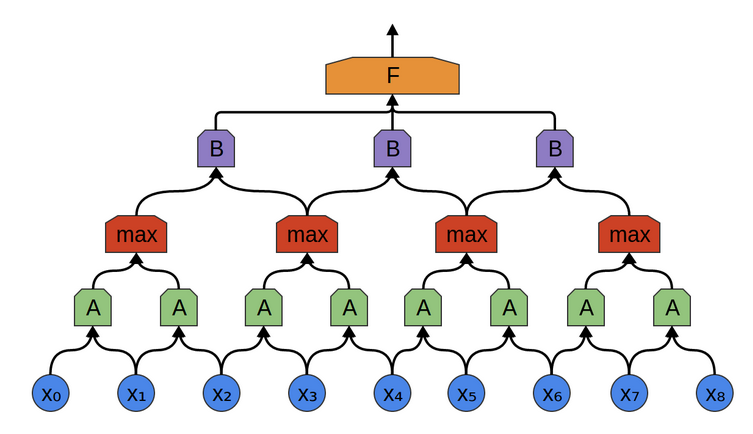} 
        \tiny
        \normalsize
    \end{minipage}
\end{figure}

The \gls{S2S reversed} uses two decoders shown in Figure \ref{fig:ed_ra}. In the first step the encoder trains together with a decoder that learns the reversed representation of the input sequence. This forces the encoder to include a backward representation of the input window in its hidden states. In a second training step, the weights of the pre-trained encoder initialize the training of the second decoder, which learns to predict the next output sequence. As shown in section \ref{seq:results} the training time is prolonged substantially by this double decoder network setup.

The second variation \gls{S2S context} uses no pre-training, instead, input information is separated into past observations of the target sequence and additional features. The encoder calculates a hidden state representation of the additional features as a sort of context. Transferring these states initializes the decoder network. The decoder uses the past target sequence as input together with the encoder weights to predict the next output sequence. This variation trains only one network and reduces computation times compared to \gls{S2S reversed}. Both networks use standard LSTM-layers within the encoder and decoder parts. 

\subsubsection{Feed Forward Neural Network}
The simplest deep learning architecture is the feed forward neural network, where information flows only in one direction (forward). The network consists of multiple layers where each node is connected to all the following nodes in the next layer. The simple structure gives this type of network an advantage in computational speed. However, its simplicity potentially results in a less complex representation of dependencies among variables reducing performance. In detail a feed forward network defines a mapping of inputs $x$ on outputs $y$ in the form of $y = f(x; \theta)$. It learns the value of the parameters $\theta$ that result in the best function approximation $f$ of $y$. In \gls{FFNN} there are no feedback connections as in recurrent neural networks like \gls{LSTM}s \citep{Goodfellow_2016}.

\subsubsection{Convolutional Long-Short Term Memory Network}
Another deep learning architecture proposed for time series prediction is the \gls{CNN}. Originally used for image processing by \cite{6795724} this architecture uses convolutional layers consisting of kernel matrices that convolve the time series information in each layer and thereby extract more complex features. This process has proven to extract important features, especially in multivariate prediction problems such as load forecasting \citep{amarasinghe_deep_2017}. 

Figure \ref{fig:cnn_layer} depicts the architecture of the \gls{CNN}-layer. Typically, time series problems use $1D$ convolutions. They mimic moving average computations and extract features across time steps and time series (spacial hierarchies). When using a $1D$ convolutional layer, a one-dimensional kernel with size $k$ functions like a weight mask that multiplied with the input layer fold inputs together to form the layer output. Often the convolutional layer output is passed on through an activation function \gls{LeakyReLU} and through a pooling layer. The pooling layer (i.e. max pooling) summarizes the feature map into a lower-dimensional representation, also called smoothing, thereby reducing the influence of small data fluctuations. The final output is flattened to either produce predictions or to be passed on to a fully connected layer. As proven by \cite{yan_multi_step_2018} the predictions benefit from adding a final \gls{LSTM} layer to predict final sequential output windows forming a \gls{CNN-LSTM} model.

\subsection{Benchmark Models}
Two prediction algorithms, one tree-based and one based on \gls{SVR} benchmark the deep learning performances. \gls{XGBoost} is a widely used and stable performing machine learning method while the \gls{MSVR} is a multioutput adaption of the \gls{SVR}. The single \gls{SVR} frequently serves as a well-performing benchmark in single output time series problems.

\subsubsection{XGBoost}
\gls{XGBoost} first proposed by \cite{ChenG16} is a classic tree-based algorithm using a gradient boosting framework. The intuition behind boosting is adding new decision trees predicting the values better where the initial model failed to give good results. Adding a multiple of these boosting trees to an ensemble of trees is expected to improve predictions by a multiple. There is no ready implemented version for multistep regression problems using the \gls{XGBoost} algorithm. This implies that one model for each time step must be fitted. 

\subsubsection{Multiple-Output Support Vector Regression}
Even though benchmarking often uses \gls{SVR}, in multistep prediction problems the basic \gls{SVR} structure restricts the algorithm to singular output values. This requires either to use predicted values iteratively as input to get the next prediction step or to fit one \gls{SVR} for each output step. In their work \cite{bao_multi-step-ahead_2014} further develop the initial version of the \gls{MSVR} first proposed by \cite{perezcruz_msvr} and design a \gls{SVR} for multistep time series prediction problems. \cite{bao_multi-step-ahead_2014} prove the \gls{MSVR} to outperform the iterative and multi-model \gls{SVR} approach in multistep time series prediction while keeping computational costs low. The \gls{MSVR} preserves the stochastic dependencies within the time series data by estimating a multidimensional output. This facilitates mapping the underlying time series dynamics by estimating a multidimensional output \citep{perezcruz_msvr}. For the full derivation of the \gls{MSVR} solution see \cite{bao_multi-step-ahead_2014}. Pre-tests of all three prediction strategies on the \gls{REFIT} data in this study confirms the effectiveness and superior performance of the \gls{MSVR} over the direct and iterative standard \gls{SVR} replicating results in \cite{bao_multi-step-ahead_2014}.

\section{Data and Experimental Design} \label{seq:exp_res}

All models are fitted to data from four homes, taken from datasets covering different geographic locations in Europe and North America. Table \ref{tab:datasets} summarizes details on the datasets and indicates whether the homes use a thermostat. The \gls{AMPds2} \citep{makelectr2016} is the largest dataset covering two full years of data. The \gls{REFIT} \citep{Murray2017AnEL} spans a period of almost two years but contains a wider gap of six weeks of missing data. \gls{GREEND} \citep{7007698} and \gls{PecanSD} \cite{7418187} are smaller datasets with data from 10 months and six months respectively with the most recent data reported from 2019 in Pecan Street. If available, the study uses the weather features published alongside the datasets. Otherwise, the preprocessing step merged historic weather data from a local weather station \footnote{obtained from https://openweathermap.org/}. The distance to the nearest weather stations is maximum 10 kilometers and the main pool of appliances selected contains appliances responsible for a large part of total appliance electric usages such as fridge, washing machine, dryer, dishwasher and television.  


{\scriptsize
\begin{xltabular}{\linewidth}{@{}  
              >{\hsize=0.75\hsize}C
             >{\hsize=1\hsize}C
             >{\hsize=1\hsize}C
             >{\hsize=0.75\hsize}C 
             >{\hsize=1.5\hsize}C 
             >{\hsize=0.75\hsize}C
             >{\hsize=1.25\hsize}C
                             @{}}
    \caption{Datasets}
    \label{tab:datasets}\\
    \toprule

    \toprule
Dataset 
& Country
  & Years
    &   WS Dist.
      &   Appliances   
        &   Thermostat 
          &  Train/Validation/Test   \\
    \midrule
\endfirsthead

    \midrule[\heavyrulewidth]
\multicolumn{7}{r}{\textit{Continue on the next page}}
\endfoot 
\endlastfoot
REFIT*
& UK, Leicester 
& 2013-11-04/2015-05-09 
& 3 km 
& fridge, dryer, washing machine, dishwasher, television
& no
& 7,540/1,884/2,807
 \\
\addlinespace
          
AMPds2
& Canada, British Columbia 
& 2012-04-01/2014-03-31 
& < 1 km 
& fridge, dryer, washing machine, dishwasher, television
& yes
& 11,710/2,927/2,883
 \\
\addlinespace
GREEND*
& Austria, Kärnten 
& 2013-12-07/2014-10-13 
& < 10 km** 
& fridge, washing machine, dishwasher, television
& n.a.
& 4,116/1,029/600
 \\
\addlinespace
PecanSD*
& USA, New York
& 2019-05-01/2019-10-31 
& < 5 km** 
& fridge, dryer, washing machine, dishwasher
& n.a.
& 2,353/588/1,447
 \\
\midrule[\heavyrulewidth]

\multicolumn{7}{l}{\scriptsize* Selected houses: GREEND - building 0; Pecan Street - house ID 3996; REFIT - house 1;} \\
\multicolumn{7}{l}{** historic weather data from a nearby weather station} \\
\normalsize
\end{xltabular}}

\subsection{Preprocessing}
The first step aggregates all original load measurements to hourly measurements and split into train, test and validation sets. Table \ref{tab:datasets} describes training, validation and test split sizes. Test data sizes vary due to testing periods starting at the beginning of a month but roughly lie between 20\% and 10\% of the full data set size. In all cases, training algorithms use 20\% of training data samples for validation.

The preprocessing pipeline imputes missing date-time values, not exceeding a gap larger than three days. The mean of the same hour on the same weekday at previous time steps imputes the missing load values to maintain presumably regular consumption patterns for weekdays. Weather variables use the mean of the corresponding week to impute values resembling values close to the same point in time. For longer time gaps this method preserves the data structure, as opposed to, i.e. forward filling introducing significantly different data patterns. The wide data gap in the \gls{REFIT} dataset was kept unimputed.  

To detect outliers, the 90\% winsorization calculates the 5th and 95th percentiles of the data and deletes all values below and above the lower and upper bounds. The subsequent step imputes the deleted measures by forward filling the previous value. In this case, forward imputation has only minor effects on the data structure and is a highly efficient method. Only fridge usage profiles in \gls{REFIT} and \gls{PecanSD} justified using outlier replacements in seven and three cases respectively. Similarly, among weather variables only wind speed in \gls{REFIT} contained extreme values. A check on historic wind speed data for reported date and location additionally verified the identified values as outliers.

Further, the Standardscaler function from sklearn standardizes features by removing the mean and scaling to unit variance with the formula $z = (x - u) / s$. Standardization of data identifies the output range beforehand and reduces the impact of larger numbers to stabilize the training process. Note that to normalize training values only the first 80\% of observations within the time series data were used to fit the Standardscaler to avoid information overspill.

After computing the features on the imputed and normalized data, rolling windows select input and output windows with 24 values (hours) comprising the training dataset. In a rolling window approach, the next data sample shifts by one value. For testing data, the experiment uses a slicing windows approach, which copies the real-world scenario where the next new 24 observations are presented all at once, not each hour. Hence the next data sample is the former sample shifted by 24 time values. Validation sets are constructed from the training set after the full preprocessing procedure.

\subsection{Model Setup}
Model input shapes after data preparation are ($N, ts, f$) with $N$ equal to the number of samples, $ts$ the number of time steps (24) and $f$ the number of features. \gls{XGBoost}, \gls{MSVR} and \gls{FFNN} require a flattened input shape, where the feature and time step dimension is flattened out to take the form ($N, ts\cdot f$). To fit input into the \gls{CNN-LSTM} model original inputs are extended to a fourth dimension to ($N, ts, f, 1$) to satisfy CNN input requirements. Most of the feature groups fit well into this shape. However, selected \gls{VEST} features and phase space transformations of the target variable change shapes to (70, 1) and (23, 2) respectively. As these transformations break up the sequential character of the input data the seq2seq architectures and the \gls{MSVR} are not considered when looking at the impact of \gls{VEST} and phase space features. The adaptations of the architecture needed to extract correct output from the adapted input dimensions make the models incomparable. 

Model architectures use the TensorFlow framework with GPU support for tuning and training. For detailed parameter settings and tuning spaces see Table \ref{tab:model_params} in Appendix \ref{app:TunParams}. All models were trained using a Google Colab Pro account with priority access to high-memory virtual machines (32 GB RAM) and GPU (T4 or P100 GPU) support. Unfortunately, as Colab resources are distributed among users, training times might vary on equal training tasks depending on the computing capacity provided.

\subsection{Evaluation Metrics}
Mainly the \gls{RMSE}, \gls{nRMSE} and a \gls{acc95} evaluate the presented results. The \gls{RMSE} measures the average magnitude of prediction errors calculated with a quadratic scoring rule that penalizes larger results more and is defined in \ref{eq:RMSE}. The \gls{nRMSE} is a scale-independent metric used by other comparable studies. It relates the \gls{RMSE} to the observed value range by the Formula in \ref{eq:nrmse}. 

The accuracy at the 95\% level \gls{acc95} (see equation \ref{eq:acc95}) calculates the sum of all predictions deviating less than 5\% from the true value and reports the share of non-deviating values in percent off all values. \cite{lachutPredictabilityEnergyUse2014} also report the \gls{acc95} as an indicator for predictability. The \gls{acc95} serves as a proxy to report the number of correctly predicted values and sets the performance indicated by the \gls{RMSE} and the \gls{nRMSE} into relation. Additionally and to ensure comparability to related literature, the results on \gls{GREEND} report the \gls{MAE}. The \gls{MAE} is defined in \ref{eq:MAE} and measures the average magnitude of prediction errors. 

\begin{equation}\label{eq:nrmse}
    nRMSE = \frac{nRMSE}{max_{y_j}-min_{y_j}}.
\end{equation}
\begin{equation}\label{eq:RMSE}
    RMSE = \sqrt{(\frac{1}{n}) \sum_{i=1}^{n} (y_j - \hat{y}_j)^2}.
\end{equation}
\begin{equation}\label{eq:acc95}
 acc95 = 
 \begin{cases}
      0 & \frac{|\hat{y}_j| - |y_j| * 100}{|y_j|}  > 5 \\
      1 & \frac{|\hat{y}_j| - |y_j| * 100}{|y_j|}  \leq 5 \\
    \end{cases}       
\end{equation}
\begin{equation}\label{eq:MAE}
    MAE = \frac{1}{n} \sum_{i=1}^{n} |y_j - \hat{y}_j|.
\end{equation}

Additionally, the \gls{MASE} defined by equation \ref{eq:MASE} reports the effectiveness of the forecasting algorithm with respect to a seasonal naïve forecaster \citep{HYNDMAN2006679}. A \gls{MASE} lower than one indicates better performance, f.e. a \gls{MASE}  of 0.5 implies that a model would have double the predictive power of a naïve forecaster. In the case of a large percentage of zero values in the data, the \gls{MASE} more reliably measures the predictive capacity of models than the combination of the \gls{nRMSE} and the \gls{acc95}.

\begin{equation} \label{eq:MASE}
    MASE = \frac{MAE}{MAE_{naive}} = 
    \frac{MAE}{\frac{1}{n} \sum_{i=1}^{n} |y_j - x_j|} 
\end{equation}

\section{Results}  \label{seq:results}
This chapter groups results in three parts. The first two subsections report on data predictability and important features. The third section elaborates on the highest performing models, prediction stability across datasets, reliable model-feature combinations and suitability of \gls{CNN-LSTM} and \gls{S2S context} models for the task at hand. All subsections will first focus on results for the fridge appliance across all data sources and second compare these results to performances on other appliance types. For appliance types except for fridge the presented results include only outcomes of \gls{REFIT}. Additionally, Table \ref{tab:fe_perform_datasets} and Figures \ref{fig:changes_model_nrmse} and \ref{fig:changes_model_acc} compare training duration across models to verify the applicability of presented approaches for customer-oriented solutions.

\subsection{Predictability} 

The \gls{wPE} measures a considerable amount of noise in the reported appliance load data similar to results in the related literature \citep{AQUINO2017277}. Further, data predictabilities of fridge profiles show higher accuracies on \gls{REFIT} in comparison to \cite{lachutPredictabilityEnergyUse2014} who reports accuracies between 50\% and 74\% a day ahead. The combination of these findings validates the \gls{wPE} as an effective indicator for predictability and verifies the effectiveness of proposed models over simpler statistical approaches in \cite{lachutPredictabilityEnergyUse2014}. The \gls{wPE} approach is suitable to pre-assess potential of prediction modeling and the amount of predictable data.

\begin{figure}
    \centering
    \caption{\\ Weighted Permutation Entropy for All Appliances and Datasets}
    \label{fig:wpe_point}
    \includegraphics[width=0.99\textwidth]{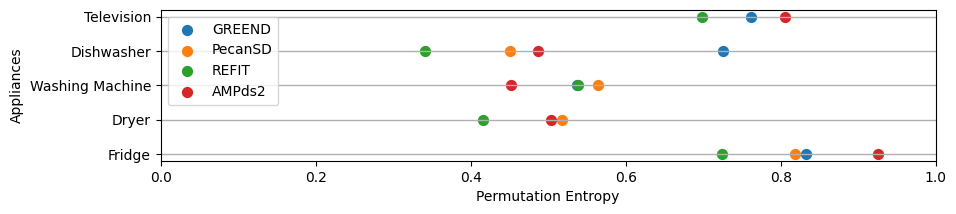}
\end{figure}

Overall the \gls{wPE} reliable assesses, which appliances and datasets are easier to predict. Models reach higher accuracies on prediction tasks specified as easier to predict and lower otherwise. Entropy values above 0.5 indicate a significant amount of randomness in the data \citep{Bandt2002PermutationEA}. According to Figure \ref{fig:wpe_point} the dishwasher and the dryer as well as fridge profiles on \gls{REFIT} show lower complexity compared to the fridge and the television and fridge profiles on the other datasets. The \gls{wPE} analysis indicates larger divergence between the television and washing machine than measured accuracies confirm. This indicates that models better fit television profiles with a higher data complexity. Further in Table \ref{tab:mase_dataset} the predictive power of models measured in terms of the \gls{MASE} values is higher for \gls{GREEND} and \gls{AMPds2} as well as for fridge and washing machine. In combination, both measures confirm that deep learning approaches perform stronger whenever data complexity is higher.

\begin{table}
\centering
\caption{MASE Values Across Datasets}
\label{tab:mase_dataset}\
\footnotesize\setlength{\tabcolsep}{1pt}
\begin{tabularx}{\textwidth} 
{
p{2 cm}
p{1.5 cm}
p{1.5 cm}
p{2 cm}
p{2 cm}
}
\\
\hline
Data &
Max &
Min &
Model Min &
FE Min 
\\
\hline

GREEND &
0.9362 & \textbf{0.6792} & LSTM & date-time
 \\
\addlinespace

PecanSD &
0.9656 & \textbf{0.7968} & S2S context & weather and date-time
 \\
\addlinespace

REFIT &
0.8755 & \textbf{0.7927} & CNN-LSTM & phase space
 \\
\addlinespace

AMPds2 &
0.9208 & \textbf{0.7057} & CNN-LSTM & date-time
 \\

\end{tabularx}
\end{table}

\begin{table}
\centering
\caption{MASE Values Across Appliances on REFIT}
\label{tab:mase_appliances}\
\footnotesize\setlength{\tabcolsep}{1pt}
\begin{tabularx}{\textwidth} 
{
p{2 cm}
p{1.5 cm}
p{1.5 cm}
p{2 cm}
p{2 cm}
}
\\
\hline
Appliance &
Max &
Min &
Model Min &
FE Min 
\\
\hline

fridge &
0.8755 & \textbf{0.7948} & CNN-LSTM & weather
 \\
\addlinespace

washing machine &
1.5525 & \textbf{0.7595} & LSTM & w + dt
 \\
\addlinespace

television &
1.3974 & \textbf{1.0758} & LSTM & w + dt
 \\
\addlinespace

dishwasher &
2.0960 & \textbf{0.8204} & LSTM & appliances
 \\
\addlinespace

dryer &
6.1995 & \textbf{1.2514} & LSTM & appliances
 \\

\end{tabularx}
\end{table}


\subsection{Important Feature Groups}
\begin{figure} 
   \centering
    \begin{minipage}{0.5\textwidth}
        \centering
        \caption{\\ Changes in Mean nRMSE Scores per Feature Group}
        \label{fig:changes_fe_nrmse}
        \includegraphics[width=0.99\textwidth]{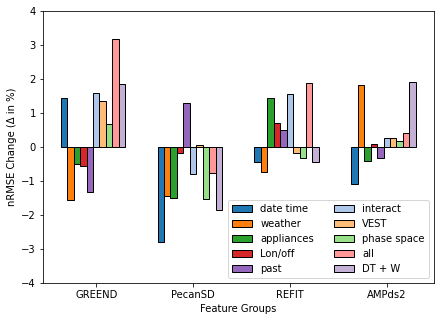}
    \end{minipage}\hfill
    \begin{minipage}{0.5\textwidth}
        \centering
        \caption{\\ Changes in Mean Accuracy per Feature Group}
        \label{fig:changes_fe_acc}
        \includegraphics[width=0.99\textwidth]{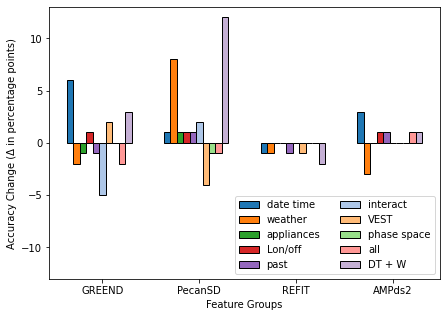} 
    \end{minipage}
\end{figure}

Figures \ref{fig:changes_fe_nrmse} and \ref{fig:changes_fe_acc} visualize the impact of features on predictions. It becomes clear that sine cosine encoded date-time features and the holiday indicator are the most important features, followed by weather features and \gls{ls-on/ls-off} indicators. The impact of engineered features remains mixed across the datasets.

Date-time features positively impact prediction quality most frequently across all datasets and provide the largest improvements compared to other feature groups. Especially in combination with weather features, models highly profit from the cyclical information added to the sequential input improving prediction accuracy and error margins. This underlines the importance of including date-time features. Both numerical encoding of date-time features and one-hot encoded time features yielded worse results in comparison to predicting the target from its past values across all datasets.

The second most important features are weather indicators, despite their less consistent contribution across datasets. Weather features perform stronger in overall error on the \gls{nRMSE} across datasets, while gains in \gls{acc95} depend on the underlying data with the risk of worsening the performance. For example, on the \gls{AMPds2} dataset performance drops significantly when including weather features. The household in the \gls{AMPds2} uses thermostats in each room. This most likely corrupts the potential information contained in outdoor environmental variables, when no thermostat is in place. Although weather features seem to be an influential factor for load predictions, their utility for predictive performance depends on other external factors, possibly not always available to application providers.

Phase space features mostly influence the overall error margin, without a clear tendency across models and datasets. However, looking closer at the results for individual model-feature combinations, interestingly, phase space features combined with \gls{CNN-LSTM} model consistently show a small improvement of the \gls{nRMSE} mostly without decreasing accuracy. This holds true across datasets and appliance types. The phase space reconstruction breaks up the sequential structure of the inputs complicating the sequential processing of values. Conclusively, the convolutional processing provides a better fit to the new data structure than sequential layer designs.    

All other feature groups show small irregular effects across datasets. Their explanatory value for the target appliance loads depend on individual household properties and are not universal explanatory. Similarly, increasing performance through engineering past, summarizing and auto-correlated features, depends on the underlying data and requires a more in-depth feature selection on individual case basis.

\subsection{Model Evaluation}\label{seq:model_results}
\begin{figure}
    \centering
    \begin{minipage}{0.5\textwidth}
        \centering
        \caption{\\ Changes in Mean nRMSE Scores per Models}
        \label{fig:changes_model_nrmse}
        \includegraphics[width=0.99\textwidth]{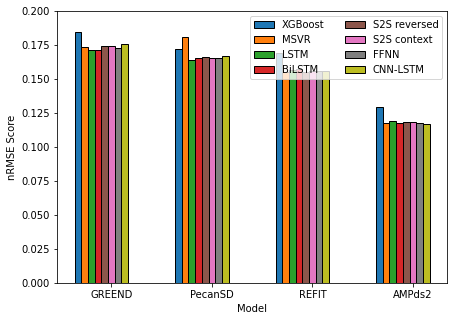}
    \end{minipage}\hfill
    \begin{minipage}{0.5\textwidth}
        \centering
        \caption{\\ Changes in Mean Accuracy per Models}
        \label{fig:changes_model_acc}
        \includegraphics[width=0.99\textwidth]{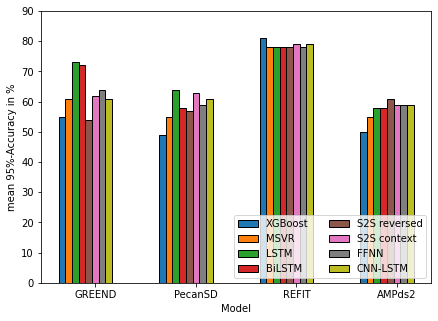} 
    \end{minipage}
\end{figure}

According to Figure \ref{fig:changes_model_nrmse} and  \ref{fig:changes_model_acc} nearly all deep learning models outperform the benchmarks in accuracy, except for \gls{REFIT} where the \gls{XGBoost} scores the highest. However, the high accuracy of the \gls{XGBoost} approach evidently requires significant concessions to higher error scores, while deep learning alternatives reach comparably high accuracy with lower error scores. This proves the consistency of deep learning performance across data sources and their good fit to complex appliance profiles.

\begin{figure}
    \centering
    \label{tab:fe_perform_datasets}
    \includegraphics[width=1\textwidth]{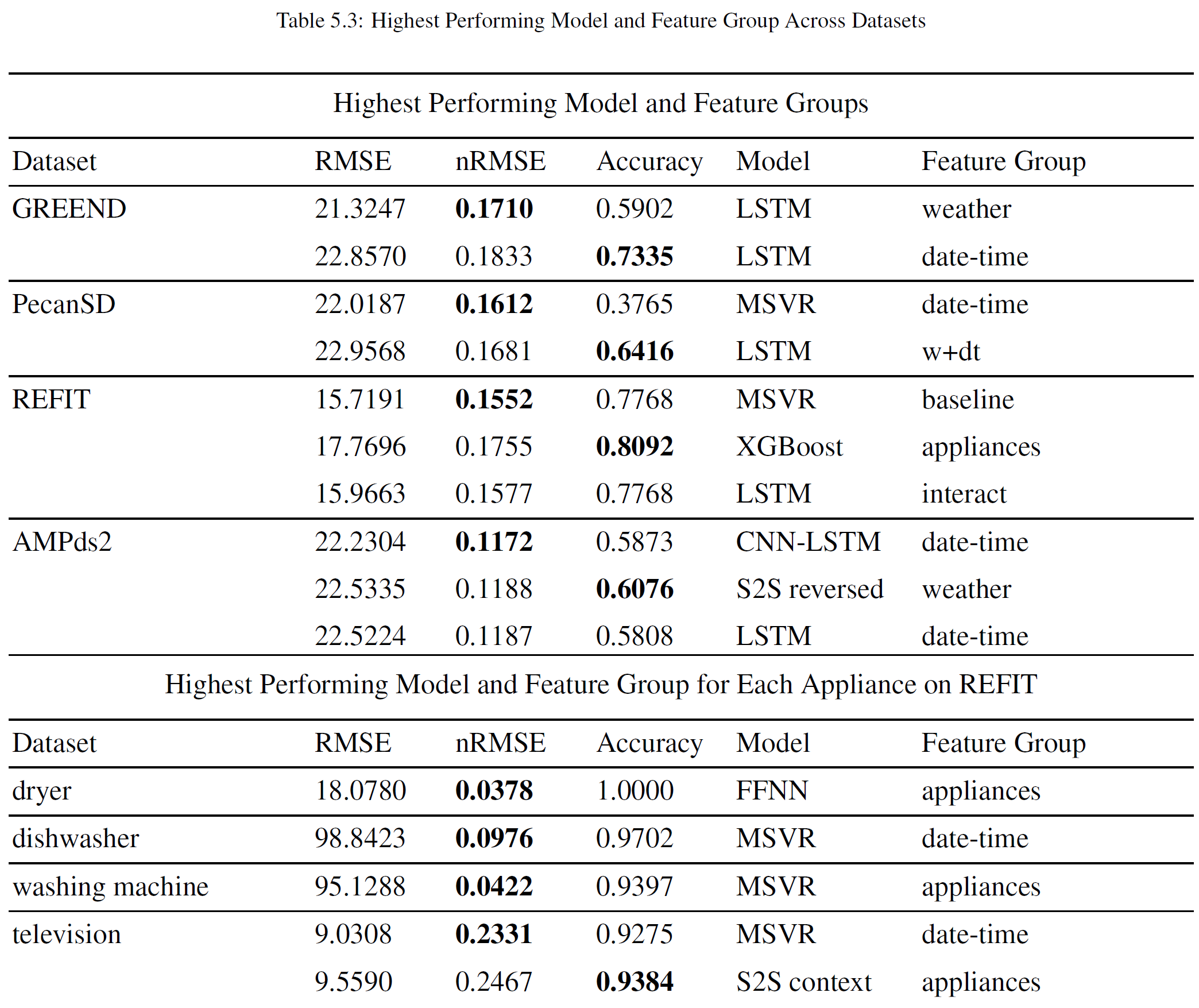}
\end{figure}

Comparing the best performing model-feature combinations referenced in Table \ref{tab:fe_perform_datasets} the \gls{LSTM} delivers high and stable performances across all datasets when combined with varying feature combinations. The \gls{LSTM} always ranks among the highest predictive performances. Especially for data with lower predictability, the \gls{LSTM} heavily draws information from date-time features. 

The \gls{S2S reversed} shows a solid performance across feature groups on average. However, the \gls{S2S reversed} is computationally expensive, resulting in longer training times This can be seen in Table \ref{tab:com_time}. Comparatively, the less complex \gls{S2S context} architecture outperforms the \gls{S2S reversed} on most of the predictions while maintaining a faster runtime (3 min 52 seconds on average). Further results indicate that the \gls{S2S context} seems to perform the best on television and \gls{AMPds2} hinting at a good fit on irregular data and appliance types. On the \gls{AMPds2}, marked as the least predictable dataset, the \gls{S2S reversed} reaches the highest performance in terms of accuracies (see Table \ref{tab:fe_perform_datasets}). 

The \gls{MSVR} generally performs very strongly in terms of overall errors alongside \gls{LSTM} and \gls{S2S reversed} especially for appliances with many zero values. Less consistent results for the \gls{MSVR} on the \gls{acc95} score show performance beneath deep learning alternatives. Lastly, the \gls{MSVR} outperforms the \gls{XGBoost} benchmark as well as the random forest algorithm\footnote{Results obtained with a random forest algorithm are not reported as exponentially growing prediction times disqualify it as a candidate for user-centric applications.}. 

\subsubsection{Performance on Different Appliance Types}

Results for other appliance types reported in Tables \ref{tab:fe_perform_datasets} and \ref{tab:mase_appliances} confirm the significance of date-time and weather as well as appliance features. When evaluating performance on other appliances it is important to keep in mind the high percentage of zero values. These appliances stay turned off most of the time with zero value percentages in the training data above 63\% for television and above 80\% for the rest in comparison to the fridge with only 25\% of zero values. High accuracies on these appliances are equally attainable by a predictor issuing zero all the time. In this case, reverting to the more reliant \gls{MASE} metric in Table \ref{tab:mase_appliances} consistently reports the \gls{LSTM} to provide the highest predictive power among algorithms and across different appliance types. The results confirm the superiority of the \gls{LSTM} model alongside weather, date-time and appliance features.

Interestingly, \gls{MASE} values for television and dryer are higher than 1 indicating that models struggle to predict better than a naive forecaster. Especially for the dryer, this result contradicts the the high predictability measured in the weighted permutation entropy. Interpreting these results, concludes that even though predictability of data is high, the models struggle to learn a proper representation, when a large share of the data contains equal values. For models to learn a representation regardless, either requires a larger ground truth database for training or specialized forecasting algorithms detecting anomalies. 

\subsubsection{Training Time}
Deployment of user friendly applications requires fast and scalable algorithms. Training duration of an algorithm on differently sized feature sets approximates both performance indicators. Table \ref{tab:com_time} reports summarized training duration of presented model architectures across different sets of features. With the second shortest duration, the \gls{LSTM} confirms its fitness as a practical solution. In general training times with deep learning (except \gls{CNN-LSTM}) vary little among different feature groups confirming their scalability to higher feature dimensions. Using \gls{S2S reversed} for production, requires higher engineering efforts for efficient computation or more powerful computation resources, as computation times are by far the highest.

\begin{table}[!htbp]
\caption{Training Time in Seconds}
\label{tab:com_time}\
\begin{tabularx}{\linewidth}{@{}  
              >{\hsize=1.55\hsize}C
             >{\hsize=1\hsize}C
             >{\hsize=0.75\hsize}C
             >{\hsize=0.75\hsize}C 
             >{\hsize=1\hsize}C 
             >{\hsize=1.1\hsize}C
             >{\hsize=1.1\hsize}C
             >{\hsize=0.75\hsize}C
             >{\hsize=1\hsize}C
                             @{}}
\hline
computation time &
XGBoost &
MSVR &
LSTM &
BiLSTM &
S2S reversed &
S2S context &
FFNN &
CNN-LSTM \\
\hline

mean & 
15 &
63 &
24 &
22 &
266 &
37 &
6 &
24
 \\

max &
66 & 
73 & 
26 & 
24 &
298 &
40 &
8 &
55
 \\

min &
8 & 
45 & 
22 &
20 &
243 &
34 &
\textbf{5} &
20
 \\
\hline

\end{tabularx}
\end{table}

\section{Discussion} \label{seq:discussion}

This section embeds reported results into related work and discusses implications and importance of results for industry and the research community. 

\cite{Ra20} demonstrate the case in which important features for multivariate appliance load prediction apply less strongly to multistep output. In their work, including \gls{ls-on/ls-off} features largely improves the reported \gls{nRMSE} of their \gls{LSTM} network (ca. +80\%). The results in this study show only a small positive impact across datasets. Importantly, the referenced authors use a different dataset\footnote{The authors use the DRED dataset.} and forecast one time step for multiple appliances. Naturally, the influence of the observed \gls{ls-on/ls-off} states 24 hours ago are not as informative as the observation of the last hour.  


The \gls{LSTM} model proves itself as a strong alternative to the \gls{S2S reversed} proposed for shorter prediction tasks in Table \ref{tab:lit_review}. On the same data, the presented \gls{LSTM} model including weather features outperforms their model on all three metrics. Subsequently in both, performance and training duration, the simpler \gls{LSTM} network is preferable over the \gls{S2S reversed}. However, the comparison should be taken carefully, as models in \cite{Razghandi2021_2} predict a 10-min resolution as opposed to one hour in this application. A more coarse forecasting granularity might be easier to predict.

The presented \gls{LSTM} outperforms the Hidden Semi Markov Model of \cite{Yuting20} that predicts the next 60 time steps of one-minute data. The differences become more evident when considering the focus of \cite{Yuting20} on groups of similar appliances including up to 50 appliances in reported target values. In previous research, single appliance loads are associated with higher error margins. Conclusively, the presented models show competitive results to the statistical approach in \cite{Yuting20}.

\subsection{Contributions}


Deep learning approaches for appliance level load prediction significantly improve with environmental and cyclical time-related information. Especially, the impact of sine cosine encoded features contradicts research assuming the sequential ordering of input values to suffice in \gls{LSTM} feature modeling and adds to the debate on efficiently encoded time features. Less complex statistical summary features, like the auto-regressive and interaction features, show lower impact and confirm the high capacity of deep learning models to autonomously extract this information from feature sets. On the other hand, chosen architectures demonstrate difficulties extracting valuable information from highly complex features, such as phase space reconstruction variables. Most likely, breaking up the sequential character of inputs disrupts efficient information extraction by models that rely on this structure. Conclusively, sequential models are no good fit for phase space reconstruction features. Correlations presented in appliance features seem of reduced importance in a multistep single appliance prediction task.

Further, the results confirm that deep learning approach are superior to alternative time series prediction methods, particularly when dealing with irregular data structures. Simpler architectures such as the \gls{LSTM} and the \gls{S2S context} showed consistently higher performance over more complex design variations such as the \gls{BiLSTM} or \gls{S2S reversed}. In conclusion, higher architectural complexity based on similar single-layer designs have only limited potential to improve predictions. Nevertheless, the potential and flexibility of new deep learning techniques and their expected development support deep learning as a trustworthy approach for modeling single appliance load profiles.

\gls{MSVR} outperforms tree-based alternatives frequently applied in related literature. This application therefore serves as a benchmark for future research work. Benchmarking with the \gls{MSVR} challenges deep learning algorithms in terms of \gls{RMSE} and \gls{nRMSE}, especially on less variable data. With a consistently low \gls{nRMSE} this model might even be a justified choice, whenever applications strongly rely on a lower error margin over higher accuracies. An example might be when high prediction errors are associated with high costs. In this case, outperforming the \gls{MSVR} indicates a good model fit to the data.

Expectations of superior performance from the \gls{CNN-LSTM} were not confirmed. Nevertheless, the model shows the best abilities to process highly complex non-sequential features such as the phase space indicators. This promising solution from aggregated load profiling specifically extracts information from input features well. Hence, the moderate performance ranking questions the importance of an additional feature extraction layer for the task at hand. 

The newly applied architecture, \gls{S2S context}, succeeds in outperforming its counterpart the \gls{S2S reversed}, with significantly less computation time. Accuracy rates fluctuate less across feature groups and indicate a lower dependence of the encoder-decoder structure on the selection of input features. Conclusively, the sequence-to-sequence structure might be preferred whenever less capacity for detailed feature engineering and selection is available.

Overall and in comparison to the existing literature on appliance profiling, transferring the feature engineering from non-deep learning approaches to the more efficient deep learning method largely improves performances. Reevaluating existing deep learning approaches for the more practical task of forecasting 24 hourly load values, contradict the proven superiority of sequence-to-sequence modeling over simpler \gls{LSTM} models in \cite{Razghandi2021_2}. Seemingly, the advantage of reverse sequence modeling diminishes with a more coarse forecasting granularity. Instead, additional information from date-time and weather features become important.

\subsection{Implications}
The presented three step framework is directly deployable in industry applications. Incorporating the presented framework would enable \gls{HEMS} and recommendation applications to transparently provide consumers with insights into their consumption profiles and showcase potential energy-savings. The predictability analysis actively manages expectations by pre-reporting expected error margins and the complexity of appliances loads chosen for predictions. The forecasting step enhanced by cyclical features and weather indicators improve the quality of existing applications and thereby comply with consumers' expectations. Further, adequately addressing consumers expectations increases satisfaction and participation, the key indicator of success in any energy-saving program \citep{SUN2017383}.

Secondly, conducting a predictability analysis prior to deployment of a modeling framework effectively assesses data potential and pre-identifies correct model structures. This speeds up application engineering and saves costs for smart home application providers, ultimately reducing entry costs of new app providers to design solutions. Further, easily accessible and low-cost engineering further nurtures innovation and development for applications tailored to consumers' needs motivating new consumers to engage in energy-saving behaviors.

\subsection{Limitations} \label{seq:limitation}
Selected deep learning architectures show restricted capacity to model all types of appliances usage profiles. Profiles of seldomly used appliances require modeling designs specialized in detecting the few positive states or larger data samples to correctly learn a profile's specificity. An interesting approach responding to a higher demand for information within these tasks extends the multistep approach by a multivariate dimension. Simultaneously modeling multiple appliance profiles in one model could ascertain whether dependencies among appliances and the prediction values of related devices provide the additional information needed to correctly predict seldom values.

The use of different geographical locations verifies the robustness of the presented results. However, the scope of this study is limited to the western hemisphere and leaves additional verification for other regions pending. Similarly, due to restrictions in available data highly promising influential features could not be tested. As an example, smart home applications might have local access to a family calendar through connected smartphone applications. The availability of this data limits a complete analysis of the predictive power of important features in appliance load modeling. Nevertheless, this work succeeds in guiding intuitions for the importance of not available features. Additionally, the predictability analysis provides fast testing for new data to form expectations for new appliance data sources.

\subsection{Recommendations}

In deployment, developers of smart home applications reliant on appliance load prediction would benefit from conducting a predictability analysis prior to implementation. This can guide suitable model choice and points to appliances requiring additional components capturing seldom behavior. A predictability step as conducted in the presented framework identifies data with a good fit to the model especially when the data structure is noisy and appliances are hardly used. In deployment this will save engineering costs and avoid promising model application performances that cannot be met.

secondly, implementation of smart meter systems and recommender applications should time label smart meter readings and connect the HEMS system to weather data either from the home itself or from a local weather station to enhance their system robustness. Accompanying options to specify household default characteristics, such as thermostat usage, delivers additional insights and prevents corrupted results from external influences. Generally advanced engineering and additional processing of input features is not needed. Rather efforts might go into more data collection of promising features not included in this study and described in the following section.

\subsection{Future Work}
Future research on multistep, multi-appliance profiling can address some of the limitations stated in Section \ref{seq:limitation}. This task comprises a higher output complexity and requires high capacity deep learning architectures such as transformers. Interestingly, transformers process complete input sequences without relying on past hidden states and sequential processing. This could preserve the long-term dependencies within time-ordered sequences and capture the cyclical characteristics important within the presented topic. Equally, mechanisms such as multi-head attention and positional embedding designed to capture the multivariate dependencies might effectively predict different appliance usage profiles simultaneously and thereby improve the difficulty in seldom used appliance profiling. 

Another interesting future path would collect additional data more directly representing human behavior. Humans are the most influential impact factor and hence occupancy, current well-being, family planners and attitudes towards smart home topics form highly interesting data sources that might further close the gap in accuracy between predictions and the ground truth. All research from how to collect this data up to its impact and predictability would significantly contribute to appliance level load applications.

\section{Conclusion}\label{seq:conclusion}
This paper examines various deep learning architectures along with important feature groups for appliance level load prediction. It shows that deep learning consistently provides more accurate predictions than tree-based and multistep \gls{SVR} benchmarks. The contributions of the paper are twofold. First, the study demonstrates the robustness of the \gls{LSTM} network across different data sets and appliance types. Secondly, it identifies cyclical encoded time features as a highly important feature group alongside weather features to enhance prediction performance. The findings additionally contribute to the distinct usage of time features in sequential forecasters demonstrating the positive impact of cyclical encoding. 

\bibliographystyle{unsrtnat}

\bibliography{sample}

\newpage
\appendix
\appendixpage
\renewcommand{\thesubsection}{\Alph{subsection}}
\renewcommand{\theequation}{\thesubsection.\arabic{equation}}
\renewcommand{\thetable}{\Alph{table}}
\renewcommand{\thetable}{\thesubsection.\arabic{table}}

\subsection{Parameters}
\subsubsection{Predictability and Feature Parameters}\label{app:wpe}


\begin{table}[!htbp]
\centering
\caption{Predictability and Feature Parameters}
\label{tab:wpe_fe_params}\
\footnotesize\setlength{\tabcolsep}{1pt}
\begin{tabularx}{\linewidth}{@{}  
              >{\hsize=1.25\hsize}C
             >{\hsize=0.75\hsize}C
             >{\hsize=0.75\hsize}C
                             @{}}
\midrule
parameters &
setting &
software (packages) \\
\midrule

order: 7 & weighted & pyentrp \\
delay: 1 & permutation &\\
normalize: False & entropy &\\
\midrule

time_delay: 1 &
Taken & giotto-tda \\
dimension: 2 & embedding & \citep{tauzin2020giottotda} \\
flatten: True & &\\


\end{tabularx}
\end{table}



\subsubsection{Model Parameters} \label{app:TunParams}

\begin{figure}
    \centering
    \label{tab:model_params}
    \includegraphics[width=1\textwidth]{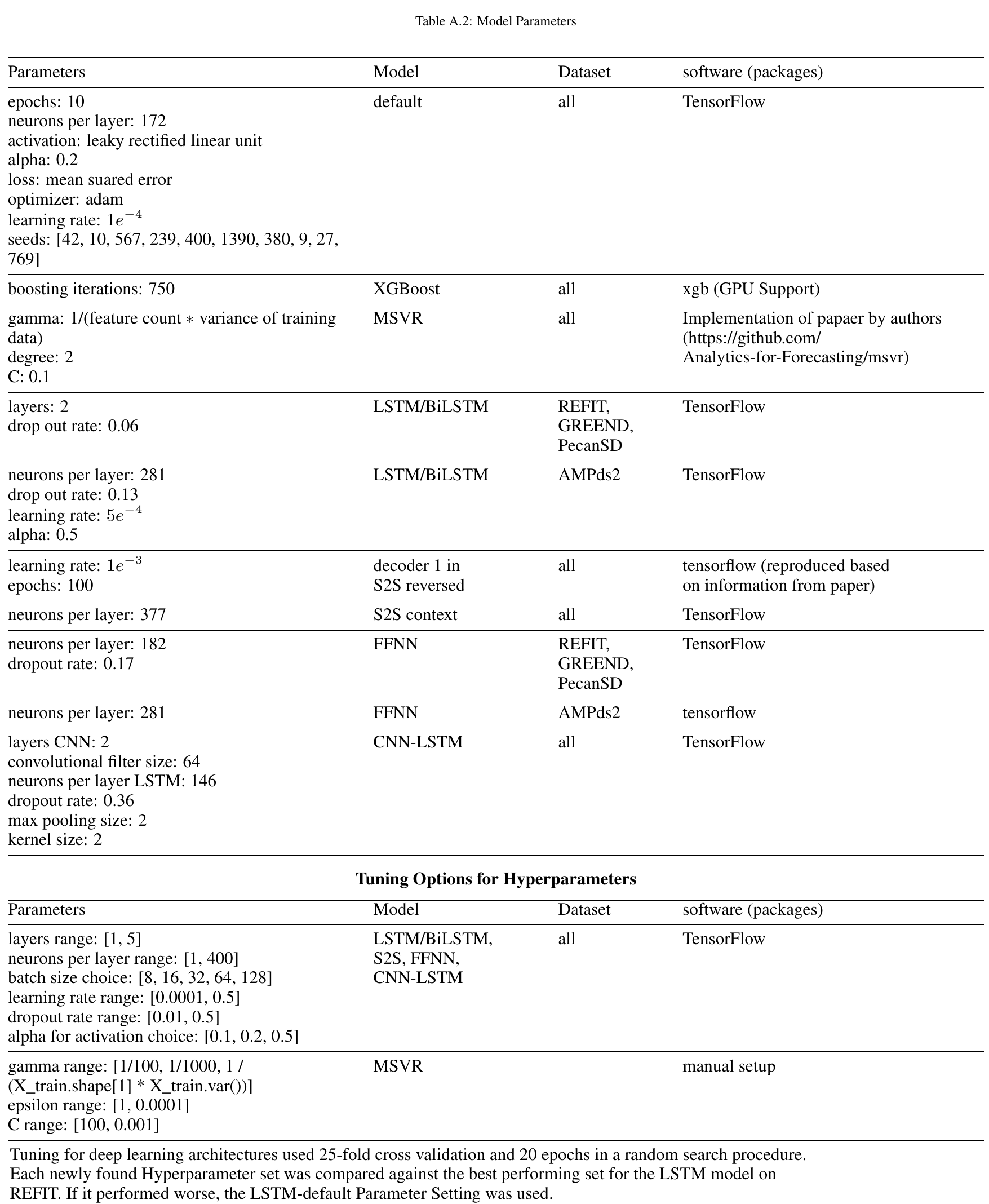}
\end{figure}

\newpage

\begin{figure}
    \centering
    \label{tab:results_fridge}
    \includegraphics[width=1\textwidth]{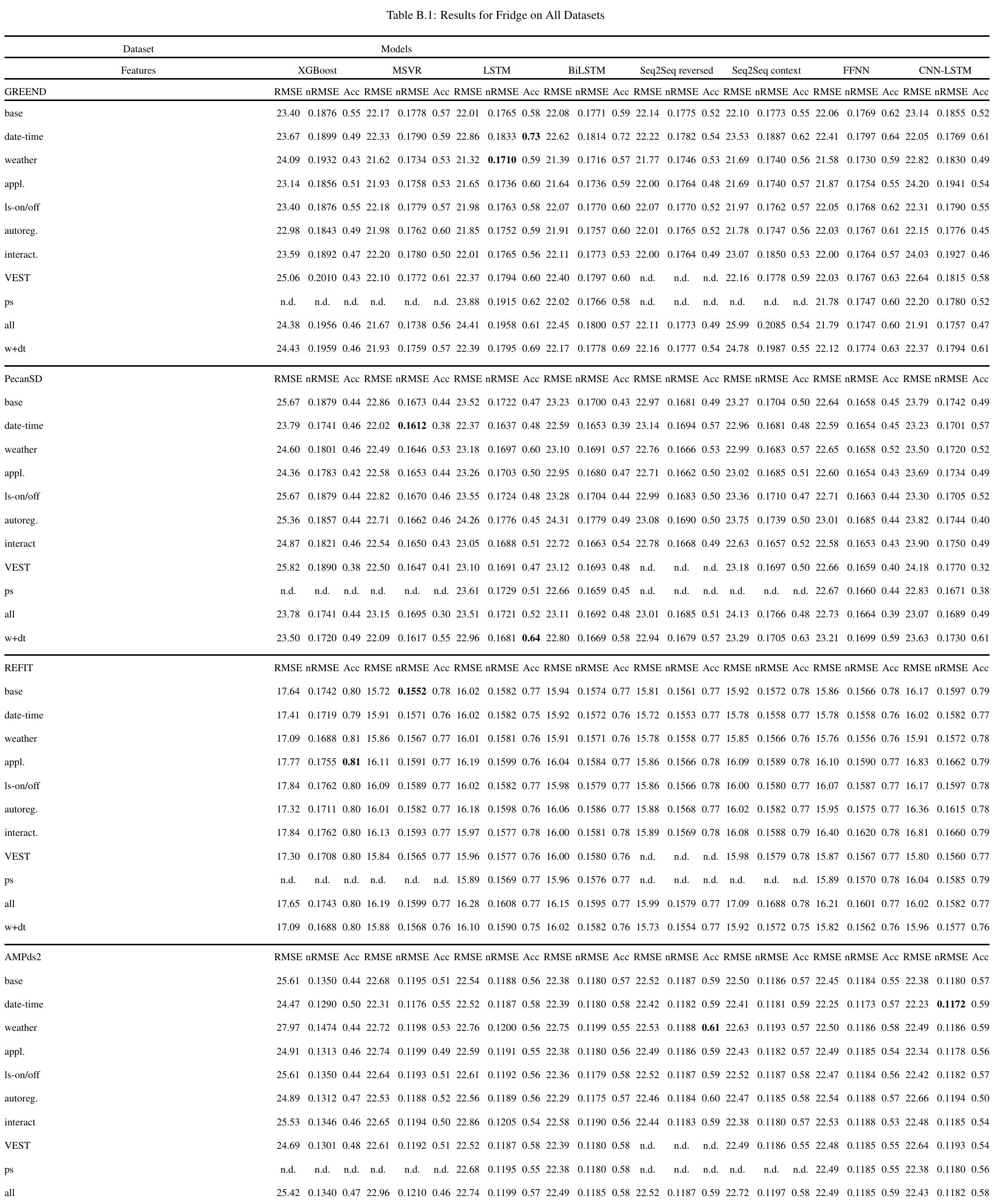}
\end{figure}

\newpage


\begin{figure}
    \centering
    \label{tab:results_appliances}
    \includegraphics[width=1\textwidth]{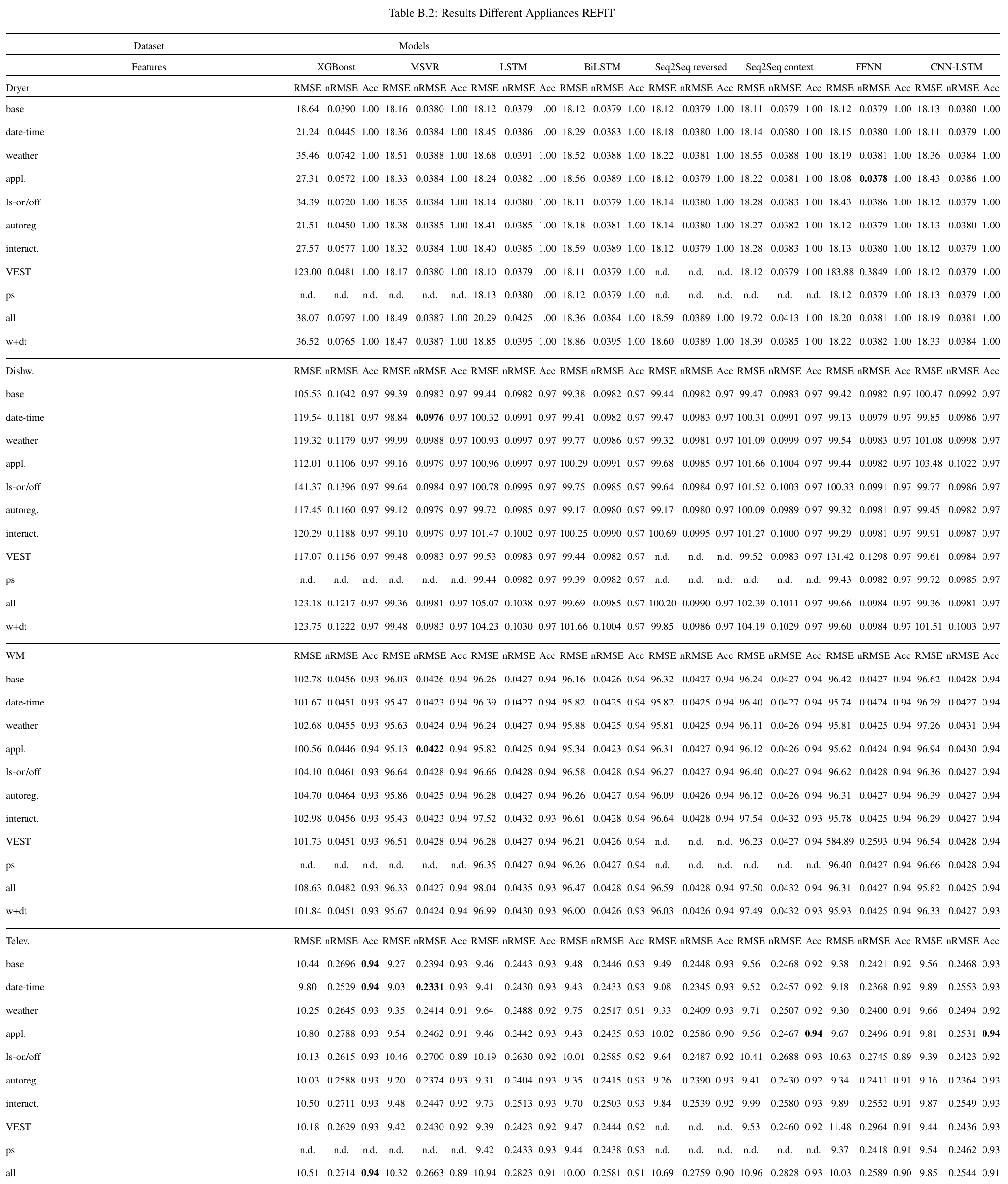}
\end{figure}

\end{document}